\documentclass{article}

\usepackage{arxiv}

\usepackage[utf8]{inputenc} 
\usepackage[T1]{fontenc}    
\usepackage{hyperref}       
\usepackage{url}            
\usepackage{booktabs}       
\usepackage{amsfonts}       
\usepackage{nicefrac}       
\usepackage{microtype}      
\usepackage{lipsum}		
\usepackage{graphicx}
\usepackage{natbib}
\usepackage{doi}

\usepackage{amsmath}
\usepackage{multirow}

\title{Enhancing Transformer-based Routing by Encoding Distance via Relative Positional Encoding}

\author{ \href{https://orcid.org/0000-0002-5534-3487}{\includegraphics[scale=0.06]{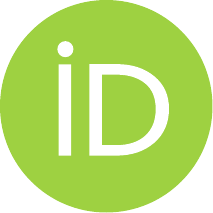}\hspace{1mm}Leyre Encío} \href{https://orcid.org/0000-0002-5746-2199}{\includegraphics[scale=0.06]{orcid.pdf}\hspace{1mm}Daniel Fuertes} \href{https://orcid.org/0000-0003-0618-3488}{\includegraphics[scale=0.06]{orcid.pdf}\hspace{1mm}Carlos R. del-Blanco} \href{https://orcid.org/0000-0001-6449-5151}{\includegraphics[scale=0.06]{orcid.pdf}\hspace{1mm}Fernando Jaureguizar} \\
	Grupo de Tratamiento de Imágenes, Information Processing and Telecomunications Center,\\
	ETSI Telecomunicación, Universidad Politécnica de Madrid, 28040, Madrid, Spain\\
	\texttt{\{leyre.encio, d.fcoiras, carlosrob.delblanco, fernando.jaureguizar\}@upm.es} \\
}

\renewcommand{\headeright}{}
\renewcommand{\undertitle}{}
\renewcommand{\shorttitle}{Enhancing Transformer-based Routing by Encoding Distance via Relative Positional Encoding}

\hypersetup{
pdftitle={Enhancing Transformer-based Routing by Encoding Distance via Relative Positional Encoding},
pdfsubject={q-bio.NC, q-bio.QM},
pdfauthor={Leyre Encío, Daniel Fuertes, Carlos R. del-Blanco, Fernando Jaureguizar},
pdfkeywords={Vehicle routing problem, deep reinforcement learning, relative positional encoding, graph transformer network},
}

\begin{document}
\maketitle

\begin{abstract}
	This paper explores Relative Positional Encoding (RPE) as an additive bias in Transformer architectures to solve the Team Orienteering Problem. By embedding in the attention mechanism pairwise spatial relationships among nodes of the graph that represents the routing problem, the transformer encoder can compute a richer spatial-aware graph embedding that allows the decoder to estimate better routes. Experimental results involving instances up to 100 nodes demonstrate consistent improvements in collected rewards and optimality gaps over vanilla Transformer architectures used by other state-of-the-art works. These findings highlight that explicit relational modeling significantly enhances scalability and generalization for complex combinatorial optimization.
\end{abstract}


\section{Introduction}
Vehicle Routing Problems (VRPs) represent an important research area in Neural Combinatorial Optimization (NCO), where a set of agents must visit a sequence of nodes under operational constraints while optimizing a given objective. These problems are inherently structured as graphs of nodes, and are typically NP-hard, making them challenging to solve exactly at scale. As a result, there has been growing interest in learning-based approaches that can exploit problem structure and generalize across instances.

The Transformer architecture, introduced by \cite{vaswani2017attention}, has become a dominant paradigm for sequence modeling due to its reliance on self-attention mechanisms instead of recurrence \cite{gama2021recurrent} or convolution \cite{qi2024convolutional}. The original Transformer computes attention as a scaled dot-product \cite{vaswani2017attention}, later extended with variants such as sparse \cite{child2019generating}, relation-aware \cite{ji2020sequential}, and linear \cite{choromanski2020rethinking} attention mechanisms. Among these, relation-aware self-attention \cite{shaw2018self} introduces pairwise positional terms, improving performance and enabling generalization beyond sequences.

\begin{figure}
    \centering
    \includegraphics[width=0.75\linewidth]{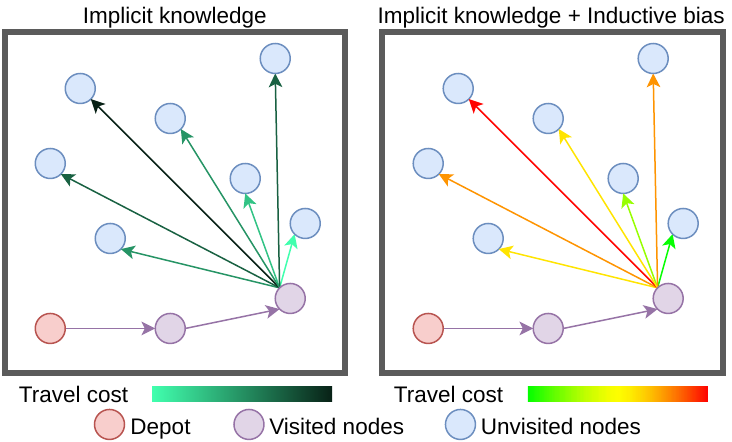}
    \caption{Example of RPE as inductive bias}
    \label{fig:RPE}
    \vspace{-0.2cm}
\end{figure}

Self-attention enables modeling long-range dependencies with parallel computation, but it is inherently permutation-invariant and thus requires explicit positional information to encode structure. Early Transformers incorporate absolute positional encodings, typically sinusoidal or learned embeddings, which are added to the input representations. While effective in natural language processing, these encodings impose a rigid notion of position that may not generalize well to graph-based problems such as VRPs. To address this limitation, relative positional encoding (RPE) was introduced, where attention scores depend on pairwise relations between elements instead of absolute indices \cite{shaw2018self}. Positional encoding methods can be grouped into: absolute encodings \cite{devlin2019bert}, which inject order but not relations; relative encodings \cite{dai2019transformer}, which model pairwise distances applied to the input embedding; and additive attention bias formulations \cite{chen2021simple,yang2025locally}, where positional information is added directly to the multi-head attention logits, allowing the model to modulate attention weights based on relative distances, which improves efficiency and scalability.

Transformers have been successfully applied using Deep Reinforcement Learning (DRL) to VRPs \cite{kool2018attention, fuertes2025, tang2026terran} using attention over node embeddings. However, many approaches rely on implicit distance encoding (absolute positional information) rather than explicitly modeling pairwise relations \cite{guan2025synergetic}, which remains underexplored.

In this work, we explore the use of RPE as an additive bias in a Transformer architecture to solve the Team Orienteering Problem (TOP) \cite{i1996team}. By explicitly embedding relational structure directly into the attention mechanism, we aim to improve generalization and solution quality compared to standard implicit distance encodings (see Fig. \ref{fig:RPE}).

\section{System description}
In this work, we have considered the TOP, which defines a complete graph $G\text{ = }(V,E)$, where each node is associated with a reward and pairwise distances. Its objective is to construct a set of routes respectively, maximizing the total collected reward under budget constraints. To address this problem, we have adopted an attention-based encoder–decoder architecture, where node features are first embedded into a latent space and subsequently processed by a stack of Transformer layers to capture global dependencies. Then, a decoder sequentially constructs feasible solutions by attending over the encoded node representations, following the standard autoregressive routing paradigm.

The core component of the proposed RPE is a modified multi-head self-attention mechanism that explicitly incorporates relational information. In the standard formulation, attention weights are computed as:

\begin{equation}
    \text{Attention}\left( Q,K,V \right) \text{= } \text{SoftMax}\left( \frac{QK^{T}}{\sqrt{d_{k}}} \right)V 
\end{equation}

which does not explicitly account for pairwise spatial structure. To address this limitation, we introduce a RPE term as an additive bias in the attention logits:

\begin{equation}
    \text{Attention}\left( Q,K,V \right) \text{= } \text{SoftMax}\left( \frac{QK^{T}}{\sqrt{d_{k}}} + B \right)V 
\end{equation}

We have constructed $B \in \mathbb{R}^{n \times n}$, where $n$ is the number of nodes, using distance information to capture complementary geometric properties of the problem. Specifically, we have considered Euclidean, Manhattan, Chebyshev distances, and cosine similarity between node coordinates, to produce a unified bias term, enabling the model to adaptively exploit different notions of proximity and directionality within the attention mechanism. The resulting architecture preserves permutation invariance while introducing a strong inductive bias aligned with the graph structure of VRPs. Moreover, since the bias is added directly to the attention logits, the proposed RPE maintains the computational efficiency and scalability of standard Transformers.

During decoding, the model constructs solutions sequentially by selecting the next node based on a probability distribution derived from the attention mechanism. At each step, infeasible actions, such as revisiting nodes or violating budget constraints, are masked to ensure valid solutions. The attention scores, augmented with the proposed bias term $B$, guide the selection process by prioritizing nodes that are both relevant in the latent space and favorable in terms of spatial relationships. This integration allows the model to jointly reason about reward, feasibility, and distance-based structure, leading to more efficient exploration of the solution space.

\section{Results}
\begin{table}[t!]
\caption{Comparison between methods for the TOP in Terms of Average Collected Reward and Gap Percentage (best in bold).} 
\label{table:TOP}
\centering
\footnotesize
\setlength{\arrayrulewidth}{0.5mm}
\begin{tabular}{ c|c|c|c|c|c|c }
 \multirow{2}{*}{Model} & \multicolumn{2}{c|}{$n=20$, $m=2$} & \multicolumn{2}{c|}{$n=50$, $m=3$} & \multicolumn{2}{c}{$n=100$, $m=5$} \\
  & Rew. & Gap & Rew. & Gap & Rew. & Gap \\
 \hline
 Gurobi & 15.810 & 0.044 & 21.547 & 44.274 & 31.535 & 62.718 \\
 ACO & 15.800 & 0.240 & 38.327 & 0.877 & 81.387 & 3.781 \\
 Vanilla & 15.779 & 0.240 & 38.285 & 0.985 & 82.788 & 2.124 \\
 Euclidean & 15.802 & 0.095 & 38.602 & 0.166 & 83.841 & 0.880 \\
 Manhattan & 15.814 & 0.019 & \textbf{38.666} & \textbf{0.000} & 83.192 & 1.647 \\
 Chebyshev & \textbf{15.817} & \textbf{0.000} & 38.572 & 0.243 & 84.139 & 0.527 \\
 Cosine & 15.801 & 0.101 & 38.573 & 0.241 & \textbf{84.585} & \textbf{0.000} \\
\end{tabular}
\end{table}

Table \ref{table:TOP} reports the performance of the proposed RPE on the TOP in terms of average collected reward and optimality gap. The comparison includes the optimization solver Gurobi \cite{Gurobi2024} and the meta-heuristic solver Ant Colony Optimization (ACO) \cite{Xiao2022aco}. It is important to note that as Gurobi takes impractically long times to generate solutions, a timeout of 60 seconds per instance has been set. All learning-based models were trained over 1280000 instances per epoch for a total of 100 epochs. Hyperparameters were kept consistent across all experiments to ensure a fair comparison.

The results show that incorporating relative positional information as an additive attention bias consistently improves the metrics over the vanilla Transformer baseline, Gurobi, and ACO across all problem sizes. This confirms that explicitly modeling pairwise relationships leads to more effective solution construction. For smaller instances ($n\text{ = }20$), where the problem is less complex and baseline methods already achieve near-optimal solutions, the improvements are relatively modest. However, as the problem size increases ($n\text{ = }50$ and $n\text{ = }100$), the benefits of the proposed approach become more pronounced. In these settings, the models incorporating RPE achieve higher rewards and lower optimality gaps, indicating a better ability to capture the underlying structure of the problem and make more informed decisions.

Overall, the results indicate that the key factor is not the specific distance metric used, but the explicit integration of relative positional information into the attention mechanism. By providing direct access to pairwise relationships, the model benefits from a stronger inductive bias, leading to improved generalization and scalability, particularly in larger and more challenging scenarios.

\section{Conclusions}

In conclusion, incorporating relative positional information as an additive attention bias significantly enhances Transformer-based solvers for the TOP. Results indicate that the RPE approach provides a superior inductive bias that improves performance and scalability as problem complexity grows. By explicitly modeling pairwise distances within the attention logits, the architecture gains a deeper understanding of the problem's spatial representation, enabling the model to predict more effective and optimized routes. This work highlights that the key to improving learning-based routing models lies in the explicit integration of relational structure. Ultimately, the proposed RPE offers a robust, scalable framework for solving graph-based NCO problems.

\section{Acknowledgements}
This work was supported in part by the Comunidad de Madrid under project TEC-2024/COM-322 (IDEALCVCM), in part by MCIU/AEI/10.13039/501100011033 of the Spanish Government under project PID2023148922OA-I00 (EEVOCATIONS), and in part by ``Ayudas a la Investigación para el Personal Docente e Investigador de la ETSIT-UPM (2026)'' under project ``SATURNO''. The authors would also like to thank Airbus Defence and Space for their support.

\bibliographystyle{unsrt}
\bibliography{references}  

@article{i1996team,
        title={The team orienteering problem},
        author={I-Ming, C and Golden, BL and Wasil, EA},
        journal={European Journal of Operational Research},
        volume={88},
        number={3},
        pages={464--474},
        year={1996},
        publisher={Elsevier Science Publishing Company, Inc.}
    }

@article{vaswani2017attention,
    title={Attention is all you need},
    author={Vaswani, Ashish and Shazeer, Noam and Parmar, Niki and Uszkoreit, Jakob and Jones, Llion and Gomez, Aidan N and Kaiser, {\L}ukasz and Polosukhin, Illia},
    journal={Advances in neural information processing systems},
    volume={30},
    year={2017}
}

@inproceedings{shaw2018self,
  title={Self-attention with relative position representations},
  author={Shaw, Peter and Uszkoreit, Jakob and Vaswani, Ashish},
  booktitle={Proceedings of the 2018 Conference of the North American Chapter of the Association for Computational Linguistics: Human Language Technologies, Volume 2 (Short Papers)},
  pages={464--468},
  year={2018}
}

@inproceedings{kool2018attention,
        title={Attention, Learn to Solve Routing Problems!},
        author={Wouter Kool and Herke van Hoof and Max Welling},
        booktitle={International Conference on Learning Representations},
        year={2019},
    }

@article{fuertes2025,
        author={Fuertes, Daniel and del-Blanco, Carlos R. and Jaureguizar, Fernando and García, Narciso},
        title={TOP-Former: A Multi-Agent Transformer Approach for the Team Orienteering Problem}, 
        journal={IEEE Transactions on Intelligent Transportation Systems}, 
        volume={26},
        number={9},
        pages={13799-13810},
        year={2025},
        doi={10.1109/TITS.2025.3566157}
    }

@article{tang2026terran,
        author={Tang, Maojie and Yu, Nanpeng and Karamouzas, Ioannis and Ye, Zuzhao},
        journal={IEEE Transactions on Automation Science and Engineering}, 
        title={TERRAN: A Transformer-Based Electric Vehicle Routing Agent for Real-Time Adaptive Navigation}, 
        year={2026},
        volume={23},
        number={},
        pages={3889-3901},
        doi={10.1109/TASE.2025.3632767}
    }

@article{guan2025synergetic,
        title={Synergetic attention-driven transformer: A Deep reinforcement learning approach for vehicle routing problems},
        author={Guan, Qingshu and Cao, Hui and Jia, Lixin and Yan, Dapeng and Chen, Badong},
        journal={Expert Systems with Applications},
        volume={274},
        pages={126961},
        year={2025},
        publisher={Elsevier}
    }

@article{gama2021recurrent,
        title = {A reinforcement learning approach to the orienteering problem with time windows},
        journal = {Computers \& Operations Research},
        volume = {133},
        pages = {105357},
        year = {2021},
        issn = {0305-0548},
        doi = {https://doi.org/10.1016/j.cor.2021.105357},
        author = {Ricardo Gama and Hugo {L. Fernandes}},
    }

@Article{qi2024convolutional,
        author = {Qi, Dingding and Zhao, Yingjun and Wang, Zhengjun and Wang, Wei and Pi, Li and Li, Longyue},
        title = {Joint Approach for Vehicle Routing Problems Based on Genetic Algorithm and Graph Convolutional Network},
        journal = {Mathematics},
        volume = {12},
        year = {2024},
        number = {19},
        issn = {2227-7390},
        doi = {10.3390/math12193144}
    }

@article{child2019generating,
    title={Generating long sequences with sparse transformers},
    author={Child, Rewon and Gray, Scott and Radford, Alec and Sutskever, Ilya},
    journal={arXiv preprint arXiv:1904.10509},
    year={2019}
}

@inproceedings{ji2020sequential,
    title={Sequential recommendation with relation-aware kernelized self-attention},
    author={Ji, Mingi and Joo, Weonyoung and Song, Kyungwoo and Kim, Yoon-Yeong and Moon, Il-Chul},
    booktitle={Proceedings of the AAAI conference on artificial intelligence},
    volume={34},
    pages={4304--4311},
    year={2020}
}

@inproceedings{choromanski2020rethinking,
    title={Rethinking Attention with Performers},
    author={Krzysztof Marcin Choromanski and Valerii Likhosherstov and David Dohan and Xingyou Song and Andreea Gane and Tamas Sarlos and Peter Hawkins and Jared Quincy Davis and Afroz Mohiuddin and Lukasz Kaiser and David Benjamin Belanger and Lucy J Colwell and Adrian Weller},
    booktitle={International Conference on Learning Representations},
    year={2021},
}

@inproceedings{devlin2019bert,
  title={Bert: Pre-training of deep bidirectional transformers for language understanding},
  author={Devlin, Jacob and Chang, Ming-Wei and Lee, Kenton and Toutanova, Kristina},
  booktitle={Proceedings of the 2019 conference of the North American chapter of the association for computational linguistics: human language technologies, volume 1 (long and short papers)},
  pages={4171--4186},
  year={2019}
}

@inproceedings{dai2019transformer,
  title={Transformer-xl: Attentive language models beyond a fixed-length context},
  author={Dai, Zihang and Yang, Zhilin and Yang, Yiming and Carbonell, Jaime G and Le, Quoc and Salakhutdinov, Ruslan},
  booktitle={Proceedings of the 57th annual meeting of the association for computational linguistics},
  pages={2978--2988},
  year={2019}
}

@inproceedings{chen2021simple,
  title={A simple and effective positional encoding for transformers},
  author={Chen, Pu-Chin and Tsai, Henry and Bhojanapalli, Srinadh and Chung, Hyung Won and Chang, Yin-Wen and Ferng, Chun-Sung},
  booktitle={Proceedings of the 2021 conference on empirical methods in natural language processing},
  pages={2974--2988},
  year={2021}
}

@article{yang2025locally,
  title={Locally enhanced denoising self-attention networks and decoupled position encoding for sequential recommendation},
  author={Yang, Xingyao and Dong, Xinsheng and Yu, Jiong and Li, Shuangquan and Xiong, Xinyu and Shen, Hongtao},
  journal={Computers and Electrical Engineering},
  volume={123},
  pages={110064},
  year={2025},
  publisher={Elsevier}
}

@inproceedings{Xiao2022aco,
        author={Xiao, Kun and Lu, Junqi and Nie, Ying and Ma, Lan and Wang, Xiangke and Wang, Guohui},
        title={A Benchmark for Multi-UAV Task Assignment of an Extended Team Orienteering Problem}, 
        booktitle={2022 China Automation Congress (CAC)}, 
        volume={},
        number={},
        pages={6966-6970},
        year={2022},
        doi={10.1109/CAC57257.2022.10054991}
    }

@misc{Gurobi2024,
        author = {{Gurobi Optimization, LLC}},
        title = {{Gurobi Optimizer Reference Manual}},
        year = 2024,
        howpublished = {\url{https://www.gurobi.com}}
    }

\end{document}